\documentclass[sigconf]{acmart}
\AtBeginDocument{%
  \providecommand\BibTeX{{%
    \normalfont B\kern-0.5em{\scshape i\kern-0.25em b}\kern-0.8em\TeX}}}

\setcopyright{acmcopyright}
\copyrightyear{2018}
\acmYear{2018}
\acmDOI{10.1145/1122445.1122456}

\acmConference[Woodstock '18]{Woodstock '18: ACM Symposium on Neural
  Gaze Detection}{June 03--05, 2018}{Woodstock, NY}
\acmBooktitle{Woodstock '18: ACM Symposium on Neural Gaze Detection,
  June 03--05, 2018, Woodstock, NY}
\acmPrice{15.00}
\acmISBN{978-1-4503-XXXX-X/18/06}

\begin{document}

\title{Toward Automated Generation of Affective Gestures from Text: \\ A Theory-Driven Approach}

%
\author{Micol Spitale}
\email{micol.spitale@polimi.it}
\orcid{https://orcid.org/0000-0002-3418-1933}
\affiliation{%
  \institution{PhD at Politecnico di Milano}
  \streetaddress{Via Golgi 39}
  \city{Milan}
  \country{Italy}
  \postcode{20133}
}
\affiliation{%
  \institution{Visiting PhD at the University of Southern California}
  \streetaddress{3710 McClintock Ave, Room 423}
  \city{Los Angeles}
  \state{California}
  \postcode{90089}
  \country{USA}}

\author{Maja J Matari\'{c}}
\email{mataric@usc.edu}
\affiliation{%
  \institution{University of Southern California}
  \streetaddress{3710 McClintock Ave, Room 423}
  \city{Los Angeles}
  \state{California}
  \postcode{90089}
  \country{USA}}

\renewcommand{\shortauthors}{Spitale et al.}

\begin{abstract}
Communication in both human-human and human-robot interaction (HRI) contexts consists of verbal (speech-based) and non-verbal (facial expressions, eye gaze, gesture, body pose, etc.) components. The verbal component contains semantic and affective information; accordingly, HRI work on the gesture component so far has focused on rule-based (mapping words to gestures) and data-driven (deep-learning) approaches to generating speech-paired gestures based on either semantics or the affective state.  Consequently, most gesture systems are confined to producing either semantically-linked or affect-based gesticures. 
This paper introduces an approach for enabling human-robot communication based on a theory-driven approach to generate speech-paired robot gestures using both semantic and affective information. Our model takes as input text and sentiment analysis, and generates robot gestures in terms of their shape, intensity, and speed. 
\end{abstract}


\maketitle

\section{Introduction and background}

Humans convey information through both verbal and nonverbal channels. About 70\% of human communication is nonverbal \cite{mehrabian2017nonverbal}, and involves body language (i.e., body posture, facial expressions) and features of speech (i.e., intonation, prosody). The communicated content contains both semantics (meaning) and affect content.

In HRI, research has explored both verbal communication \cite{steels2003evolving, bisk2016natural, kennedy2017child} leveraging advances in Natural Language Processing (NLP) \cite{tellex2020robots}, and nonverbal communication (e.g., prosody \cite{marchi2014voice, saunders2011towards, sadouohi2017creating}, robot gesture \cite{xiao2014human, bremner2009conversational}, facial expressions \cite{deng2019cgan, ge2008facial}, and gaze \cite{admoni2017social, moon2014meet}). Work HCI, simulated agents, and virtual humans has also explored simulated nonverbal communication features \cite{ruhland2015review,norouzi2018systematic}.


\begin{figure*}[h]
    \centering
    \includegraphics[width=0.8\linewidth]{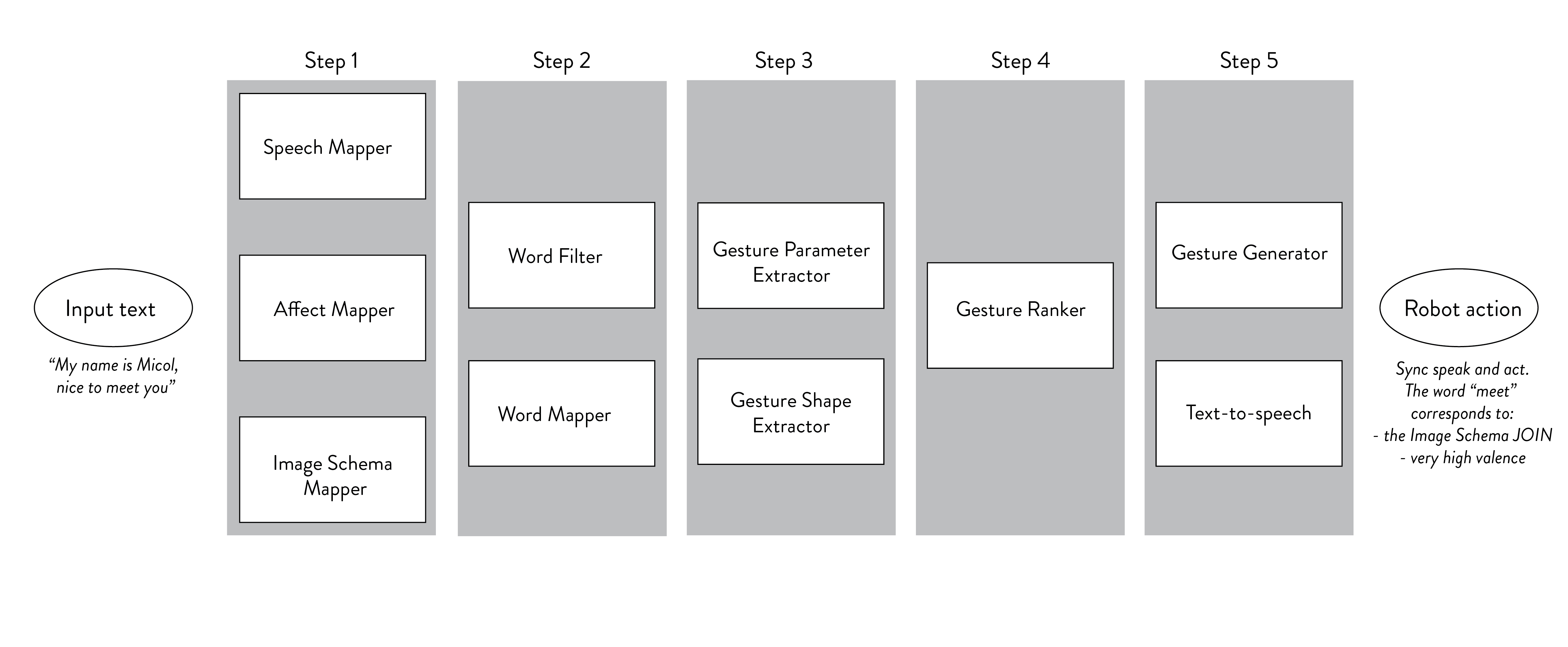}
    \caption{Process for automatically generating iconic affective robot gestures.}
    \label{fig:my_label}
\end{figure*}

Synchronizing speech and gesture is an important component of natural and effective robot communication in HRI. Work on automated gesture generation to date has focused on rule-based approaches (i.e., creating dictionaries for mapping words to gestures \cite{huang2013modeling}) and data-driven approaches (i.e., exploiting deep-learning \cite{yoon2019robots, kucherenko2020gesticulator}), producing speech-paired gestures on either semantics or affective state, respectively.  A few execeptions, including \citet{lhommet2013gesture} and \citet{ravenet2018automating}, designed rule-based models to produce both semantically-linked and affective-based gestures, but that work was only done for virtual agents and not robots. Rule-based approaches are limited to fixed {\it a priori} mappings, while data-driven approaches lack interpretability.

This work introduces the design of a theory-driven approach for automatically generating speech-paired robot gestures based on both semantic and affective information. Our design is grounded in the theories of Image Schema \cite{johnson1987body, george1987women, rohrer2005image} and Embodied Cognition \cite{wilson2002six}. \citet{johnson1987body} defined an {\it image  schema}  as  "a  recurring  dynamic pattern  of  our  perceptual interactions and motor programs that gives coherence and structure to our experience". \citet{wilson2002six} defined {\it embodied cognition} on the theory that "cognitive processes are rooted in the body’s interactions with the world". Additionally, our approach is informed by \citet{mcneill1992hand}'s classification of gestures as iconic (i.e., resembling a physical aspect of conveyed information), metaphoric (i.e., iconic gestures that refer to abstract concepts instead of concrete entities), deictic (i.e., pointing to an object, person, or directions), and beat (i.e., dictated by the rhythm of speech).

Our theory-driven approach is designed for HRI and consists of a sequence of modules with a single goal. The model receives text as input, analyzes its semantic and affective dimensions, and then generates speech-synchronized gestures that communicate both meaning and emotion. 

This work aims to make the following contributions:
\begin{enumerate}
    \item The first use of a combination of theories of Image Schema and Embodied Cognition;
    \item The ability to automatically generate gestures that communicate both semantic and affective components of speech;
    \item The ability to use machine learning as part of the gesture generation in an interpretable way.
\end{enumerate}

\section{From theories to a computational model}

From the psychology literature, the embodied cognition theory refers to the idea that “states of the body modify states of the mind” \cite{wilson2013embodied}. As part of our reasoning, we interact with the world through embodiment via language structures (acoustical and optical) which results in the mental representations we exploited to inform our abstract and concrete concepts.
\citet{johnson1987body} suggested that as humans we use similar patterns of reasoning, called Image Schemas. Image Schema enables us to map common concepts to different concrete and abstract entities. It enhances our perception and explains how we interact and move in a specific environment, grounding the language structure and meaning.   
Within the cognitive linguistic literature on embodied cognition, \citet{cienki2013image} suggested that the language structure and meaning is based on the embodied experience, and that they are naturally aligned with the production of gestures during speech. Many authors showed promising results about the use of Image Schema as mapping between conceptual entity and the language production for both speech and gestures \cite{lucking2016finding, chui2011conceptual}. 

\citet{mittelberg2018gestures} described the gesture of  mimicking the shape of a box, and the authors explained how this gesture can represent the Image Schema OBJECT or CONTAINER. The CONTAINER recalls the idea of having an entity within boundaries which elements can be inside or outside it. 
Each Image Schema leads to a different representation in the real world characterized by different body position and movements.

We took inspiration from these works to design our new model which is grounded on Embodied Cognition and Image schema theories, and which sought to bridge the acoustic (i.e., speech) and optical (i.e., gesture) language structures to inform robotic platforms.

\section{Model Architecture}

Our theory-based approach involved a sequence of modules that generate a gesture synchronized with speech from a text input, as follows:
\begin{enumerate}
    \item \textit{{Text analysis.}} 
    Speech Mapper, Affect Mapper, and Image Schema Mapper modules run in parallel and analyze the timing, sentiment, and semantics of the text, respectively. Speech Mapper returns the timing and duration of all the words in the text, the Affect Mapper maps the words to an affective meaning that corresponds to the valence, arousal, and dominance values obtained from affective lexicon analysis \cite{vad-acl2018}. In case the input text has no words that correspond to any affective value, we compute the sentiment of the whole sentence \cite{medhat2014sentiment}. Next, the Image Schema Mapper returns the corresponding Image Schema of the words that represent an iconic meaning in the input text. 
    \item \textit{{Word filtering and mapping.}} The Word Filter analyzes the input text and tags part of speech exploiting an NLU library \cite{hardeniya2016natural}. In addition, it discards words that do not provide any semantic value (e.g., prepositions, articles). The Word Mapper returns only words with a corresponding Image Schema mappings and emotional states, as well as speech-timing information. If a word associated with an Image Schema does not correspond to any affective state, it is associated with the sentiment of the whole sentence. 
    \item \textit{{Extrapolating gesture shape and parameters.}} The Gesture Parameter Extractor takes the affective state of each word and it returns the associated gesture speech, timing and amplitude according to valence and arousal value of the corresponding word \cite{van2018generic}. The Gesture Shape Extractor takes as input the Image Schema corresponding to the word and it returns the shape of the gesture.
    \item \textit{{Gesture selection.}} The Gesture Ranker makes decisions on what gesture to perform. This module ranks gestures based on their affective state and meaning in the context of the sentence. The results from the previous steps could be two or more gestures associated to specific words with iconic meaning. To ensure that it is acting naturally and believably, we rank the gestures in the sentence so that the robot performs a single gesture per sentence.
    \item \textit{{Gesture synthesis.}} The model synthesizes the text into speech (Text-to-Speech), and generates the gesture (Gesture Generator) selected in previous steps (e.g., motion of robot joints according to the input text, synchronized with speech) exploiting the MoveIt framework \cite{coleman2014reducing}, also making it ROS-compatible.
\end{enumerate}

Our ongoing work on the model involves an empirical study that evaluates the model's effectiveness and interpretability.

\section{Conclusion and Future Work}
This work introduced an approach for automatically generating robot gestures from speech with both semantic and affective meaning. To the best of our knowledge, this is the first gesture generator grounded in cognitive linguistic theories and communicates both semantic and affective information. The interpretable approach uses a sequence of single-purpose modules that exploit both data-driven and rule-based methods.
In ongoing work, we are evaluating the approach empirically to assess its efficacy and interpretability.

\begin{acks}
This research was supported  in part by the University of Southern California (supporting Maja Matari\'c), and in part by EIT Digital (supporting Micol Spitale).

The authors thank Gianna Beck and Sarah Okamoto for their help with the implementation process.
\end{acks}

\bibliographystyle{ACM-Reference-Format}
\bibliography{main}


\end{document}